\documentclass[11pt]{article}

\usepackage[preprint]{acl}

\usepackage{times}
\usepackage{latexsym}

\usepackage[T1]{fontenc}

\usepackage[utf8]{inputenc}

\usepackage{microtype}

\usepackage{inconsolata}

\usepackage{graphicx}

\usepackage{algorithm}
\usepackage{algorithmic}
\usepackage{enumitem}
\usepackage{booktabs}

\usepackage{amsmath}
\usepackage{tcolorbox}

\usepackage{tcolorbox}
\usepackage{float}
\newtcolorbox{promptbox}[1]{
  colback=gray!5, colframe=gray!50,
  coltitle=black,              
  title=#1, fonttitle=\bfseries\small,
  boxrule=0.5pt, arc=2pt
}

\usepackage{caption}
\DeclareCaptionType{prompt}[Prompt][List of Prompts]

\usepackage{stfloats}
\usepackage{multirow}

\title{Similar but Not Redundant: Embedding-Gated Refinement for High-Granularity Topic Models}

\author{
  \textbf{Daniel Denley\textsuperscript{1,2}},
  \textbf{Robert L. Nicol\textsuperscript{2}},
  \textbf{Mike J. Chantler\textsuperscript{1}}
\\
  \textsuperscript{1}Strategic Futures Lab, Heriot-Watt University, Edinburgh, United Kingdom
\\
  \textsuperscript{2}STMicroelectronics, Edinburgh, United Kingdom
\\
  \small{\texttt{dad2001@hw.ac.uk}, \texttt{robert.nicol@st.com}, \texttt{M.J.Chantler@hw.ac.uk}}
}

\begin{document}
\maketitle
\begin{abstract}
High-granularity topic modelling extracts hundreds of niche topics from a corpus, yet this regime is rarely studied and is more commonly framed as a failure mode: "over-clustering," in which coherent themes fragment into redundant subtopics. We argue that this framing conflates genuine redundancy with topics that are semantically similar but meaningfully distinct, and that the central challenge in high-granularity modelling is separating the two. Recent post-hoc refinement methods address topic redundancy through either embedding-based similarity thresholding or LLM-guided merging. We show that neither suffices in high-granularity settings and propose embedding-gated iterative merge (EGIM), a hybrid approach that iteratively generates context-rich topic representations, uses embedding similarity to propose candidate merges across the topic space, then applies LLM judgement to accept only genuinely redundant pairs while preserving niche distinctions. Across two patent corpora and one legal corpus, LLM judgement rejects 83\% of embedding-identified candidates as non-redundant. EGIM improves discriminability on merged topics by 25–55\% over a prior iterative LLM baseline while better preserving the original topic distribution.
\end{abstract}

\section{Introduction}

Standard topic model benchmarks operate at moderate granularity, typically between 20 and 100 topics. Yet a range of downstream text-mining tasks, such as competitive intelligence in patent informatics \cite{patentminer} and area-of-law mapping in legal corpora \cite{legalexample}, can benefit from niche topic extraction via high-granularity topic model regimes. When models are pushed to higher topic counts, the result is commonly framed as "over-clustering": outputs that are overly complex and yield many redundant or highly similar topics, reducing topic interpretability and practical usefulness \cite{greene2014, meaney2023, gan2021}. We argue that this framing conflates genuinely redundant topics with semantically similar but distinct ones, and that the core challenge in high-granularity topic modelling is developing refinement methods that can distinguish the two.

Topic redundancy is a well-documented issue in topic modelling \cite{greene2014, ocallaghan2015}, including with LLM-generated topic descriptors \cite{mu2024,singh2025}. Prior work on post-hoc reduction of topic redundancy falls into two broad categories: similarity-threshold methods, which merge topics based on vector distances \cite{grootendorst2022}, and LLM-guided methods, which prompt language models to identify and consolidate redundant topics \cite{janssens2025, pham2024}. We show that neither is adequate in high-granularity regimes: threshold methods cannot distinguish genuinely redundant pairs from semantically similar ones, and LLM-guided methods are susceptible to \emph{semantic snowballing}, in which progressively broader merged labels attract further merges and collapse the topic distribution.

We propose \textbf{embedding-gated iterative merge (EGIM)}, a hybrid post-hoc refinement method that uses embedding similarity to select candidate topic pairs, then reduces each decision to a binary merge-or-keep judgment by a single LLM call. Because each call concerns one pair, the prompt can include representative documents from both topics, grounding the decision in corpus evidence rather than label semantics alone. Candidate pairs are selected via k-means sampling across the topic-embedding space rather than by greedy similarity sampling; we show in an ablation that this coverage-based strategy identifies more true duplicates. Because EGIM operates as a merge-keep loop, termination is conditioned on convergence (a fixed number of consecutive iterations without a merge) rather than a predefined target topic count. In practice, merge occurrences decrease steadily across iterations and the method converges naturally.

To evaluate post-hoc refinement in high-granularity regimes, we introduce two diagnostics: \emph{topic dispersion}, which measures distributional balance and penalises mega-topics more heavily than niche ones, and \emph{topic discriminability}, which measures how well a topic's representation aligns with its assigned documents relative to non-assigned documents. Across two patent corpora and a legal corpus, we find that 83\% of high-similarity topic pairs are rejected as non-redundant by the LLM. EGIM improves topic discriminability by 25--55\% over an iterative LLM baseline while preserving over 94\% of original topic dispersion across all three corpora, compared to 34--88\% for the baseline.

Our main contributions are:
\begin{itemize}
    \item \textbf{Embedding-gated iterative refinement.} We propose
    EGIM, a hybrid post-hoc refinement algorithm that selects candidate
    pairs via embedding similarity with coverage-based sampling, then
    applies LLM judgement grounded in representative documents to merge
    genuine duplicates whilst preserving semantically similar but
    distinct topics. The method is convergence-terminated and does not
    require a predefined target topic count.
    \item \textbf{Evaluation diagnostics for high-granularity
    refinement.} We propose topic dispersion and topic discriminability
    as diagnostics designed for high topic counts, replacing standard
    metrics that treat inter-topic similarity as sufficient evidence of
    redundancy.
\end{itemize}

\section{Related Work}

\paragraph{High-Granularity Topic Modeling.}
Most topic modeling research evaluates at low to moderate granularity, with topic counts optimized for automated metrics or drawn from standardized ranges on benchmark corpora. Work explicitly targeting high-granularity regimes is typically driven by downstream requirements. \citet{boutaleb2024} propose BERTrend, a framework for detecting weak thematic signals and emerging trends, built on BERTopic with hyperparameters selected for the explicit purpose of producing fine-grained topics. The FIDELITY framework \citep{singh2025} identifies a research gap in topic modeling approaches for high-granularity output and proposes clustering extracted keyphrases to produce fine-grained phrasal topics. \citet{singh2025} also raise concerns about their post-hoc redundancy method based on cosine similarity, noting it may oversimplify distinctions between closely related topics, and suggest LLM-based merging as a potential remedy.

\paragraph{Post-hoc Topic Refinement.}
Existing approaches to post-hoc topic redundancy reduction fall into two broad categories.

Threshold-based strategies consolidate topics by the similarity of their vector representations. BERTopic \citep{grootendorst2022} implements topic reduction via agglomerative clustering over class-based TF-IDF representations. \citet{blair2020} apply a comparable strategy to aggregated LDA ensembles, merging topics that exceed predefined cosine similarity and Jensen--Shannon divergence thresholds. These methods are efficient but cannot judge whether two similar topics are genuinely redundant or capture meaningful distinctions at fine granularity.

LLM-facilitated methods replace or supplement similarity thresholds with language model judgment. PromptTopic \citep{wang2023} includes a topic-collapse stage that uses an LLM to merge overlapping topics based solely on topic names until a target topic count is obtained. TopicGPT \citep{pham2024} similarly details a single-step hybrid clean-up stage in which they first identify topic labels above a set threshold (>0.5 cosine similarity) before passing topic lists to an LLM to make merge decisions. Most similar to our work is that of \citet{janssens2025}, who specifically investigates reduction of overlapping topics from base model outputs (BERTopic). \citet{janssens2025} iteratively utilises an LLM to perform both topic candidate selection and merged topic representation from the current topic representation list, until a desired topic count is reached.

Our work extends this line of research by decoupling candidate selection from merge judgment. Prior LLM-facilitated methods present the model with full or partial topic lists, requiring it to simultaneously identify merge candidates and execute merges. This scales poorly to high-granularity settings, where hundreds of thematically narrow topics exhaust the context window and leave little room for the document-level evidence needed to distinguish genuine redundancy from meaningful specificity. Our method instead uses semantic embeddings for thematically-aware candidate selection, reducing each merge decision to a single topic pair presented alongside representative documents for binary LLM judgment. The topic landscape is then updated and the process repeats, allowing natural convergence rather than reduction to a predefined count. Empirically, we find that the LLM rejects the majority of embedding-selected candidates, particularly in later iterations, demonstrating that similarity-threshold selection alone systematically conflates semantic proximity with genuine redundancy at high granularity.

\section{Methodology}
\subsection{Embedding-Gated Iterative Merge}
\label{sec:method}
Given an initial high-granularity topic model with document--topic assignments and a keyword-based topic representation, we first generate contextually rich topic representations by prompting an LLM with each topic's keywords and representative documents to produce a descriptive label and summary. The refinement algorithm then operates over these LLM-generated representations, iteratively removing redundant topics. The algorithm does not require specifying a target topic count; it
terminates when no merge occurs for $N$ consecutive rounds (patience
parameter) or when an LLM call budget is reached. The output is an updated topic set with revised representations and merged assignments.
\begin{algorithm}[t]
\caption{Embedding-Gated Iterative Merge (EGIM)}
\label{alg:egim}
\begin{algorithmic}[1]
\REQUIRE Topic representations $\{r_t\}$, representative documents $\{S_t\}$, similarity threshold $\tau$, patience $N$, target cluster size $s$, excluded pair cache $\mathcal{E} \leftarrow \emptyset$
\WHILE{patience $< N$}
  \STATE $z_t \leftarrow \textsc{Embed}(r_t)$ for all $t$
  \STATE $M \leftarrow \max(2,\lfloor K/s\rfloor)$ where $K = |\{z_t\}|$
  \STATE $\mathcal{C} \leftarrow \textsc{KMeans}(\{z_t\}, M)$
  \STATE $\mathcal{P} \leftarrow$ highest-sim.\ pair per cluster where $\cos(z_i,z_j)\ge\tau$
  \STATE $\mathcal{P} \leftarrow \mathcal{P} \setminus \mathcal{E}$ \COMMENT{skip excluded pairs}
  \STATE merged $\leftarrow$ \textbf{false}
  \FOR{each $(i,j)\in\mathcal{P}$}
     \STATE action $\leftarrow \textsc{LLM}(r_i,S_i,r_j,S_j)$
     \IF{action $=$ \textsc{Merge}}
        \STATE Combine topics; generate new representation
        \STATE merged $\leftarrow$ \textbf{true}
     \ELSE
        \STATE $\mathcal{E} \leftarrow \mathcal{E} \cup \{(i,j)\}$ \COMMENT{cache}
     \ENDIF
  \ENDFOR
  \IF{merged}
    \STATE patience $\leftarrow 0$
  \ELSE
    \STATE patience $\leftarrow$ patience $+ 1$
  \ENDIF
\ENDWHILE
\end{algorithmic}
\end{algorithm}

At each iteration, topic embeddings are recomputed to reflect any merges from the previous round, then clustered into $M{=}\max(2,\lfloor K/s\rfloor)$ groups via k-means, where $K$ is the current number of active topics and $s$ is a target cluster size. This makes $M$ adaptive: as topics are merged and $K$ decreases, fewer clusters are formed.

The highest-cosine-similarity unreviewed pair above threshold $\tau$ is selected from each cluster, ensuring broad coverage across the embedding space rather than concentrating on a single high-density region. In practice, this yields more successful merges per LLM call than greedy selection over the full space.

Representative documents $S_t$ are selected to balance relevance and diversity via MMR-based selection with quality filtering. When retrieving representative documents, if the topic is a product of prior merges, representative documents are drawn as evenly as possible from each underlying topic to maintain balanced coverage of the merged topic's thematic scope.

For each candidate pair, we provide the LLM with both topic representations and representative documents and ask it to choose one of two actions:

\begin{samepage}
\begin{itemize}
    \item \textbf{Merge:} combine the two topics, producing
    a new representation and merged assignment set.
    \item \textbf{Keep:} leave both topics unchanged when they
    are similar but represent meaningfully distinct concepts.
\end{itemize}
\end{samepage}

The prompts encode a conservative granularity policy that favours \emph{Keep} when topics address distinct aspects of a shared domain, and \emph{Merge} only when they describe the same underlying concept; in practice, this policy can be adjusted to reflect application needs. We cache declined pairs ($\mathcal{E}$) to avoid repeated judgement calls; since merges produce a new topic, cached entries involving the constituent topics naturally become stale and the new topic is evaluated afresh against its neighbours. Full prompt templates and implementation details are provided in Appendix~\ref{app:prompts} and Appendix~\ref{app:egim-details}.

\subsection{Datasets}

We evaluate on three corpora from domains where fine-grained thematic
distinctions carry practical value: two patent collections obtained via
Lens.org\footnote{\url{https://www.lens.org/}} \citep{jefferson2021lens}
and one legal case reports corpus.
\begin{itemize}[nosep,leftmargin=*]
\item \textbf{SPAD Patents} (1,102 documents): Patents related to
Single-Photon Avalanche Diodes, a niche semiconductor domain.
\item \textbf{Laser Defense Patents} (3,302 documents): Laser technology
patents assigned to defense organizations.
\item \textbf{Legal Case Reports} (3,890 documents): Australian Federal
Court cases full text dataset \citep{legal_case_reports}, taken from the UCI Machine Learning Repository \footnote{\url{https://archive.ics.uci.edu/dataset/239/legal+case+reports}}.
\end{itemize}
\noindent Patents are represented using description section text. Full curation
details, including pre-filtering criteria, are provided in
Appendix~\ref{app:datasets}.

\subsection{Experimental Design}

We split our evaluation into two stages. Stage~A performs a multi-factor calibration sweep to characterize how modeling choices interact in high-granularity topic modeling pipelines. In Stage~B, we apply the insights from Stage~A to produce high-granularity topic models on our datasets, then isolate the effect of our post-hoc refinement.

\paragraph{Base Topic Model.}
We adopt BERTopic \citep{grootendorst2022} as our base topic model to produce initial document--topic assignments and topic representations. BERTopic is representative of the embedding-cluster framework popular in modern topic modeling: documents or document chunks are encoded with a sentence-transformer model, embeddings reduced via UMAP \cite{umap}, clustered with HDBSCAN \cite{hdbscan}, and each cluster represented via class-based term frequency--inverse document frequency (c-TF-IDF) keyword extraction.

\paragraph{Stage~A: Base Model Configuration Sweep.}
Stage~A investigates how chunk size and preprocessing interact with granularity in the base BERTopic model. We perform a factorial sweep over \emph{corpus} $\times$ \emph{preprocessing} $\times$ \emph{chunk size} $\times$ $K$: three corpora, six preprocessing variants, three chunk sizes $\{2, 5, 10\}$~sentences, and five target granularities $K \in \{20, 40, 80, 160, 320\}$, yielding $270$ total configurations. As BERTopic's component models (UMAP and HDBSCAN) are sensitive to the underlying data distribution and desired cluster granularity, we additionally optimize each configuration for 50~trials using Optuna~\citep{optuna}, with a composite objective balancing topic quality and cluster validity (Appendix~\ref{app:optimisation}).

\paragraph{Stage~B: Post-hoc Refinement Comparison.}
Stage~B isolates the effect of refinement by fixing chunk size to 2~sentences and target granularity to $K{=}320$, guided by Stage~A findings (Section~\ref{sec:stageA-results}), while retaining all six preprocessing variants across all three corpora to give 18~base topic models. Each refinement method is applied to every base model and compared using dispersion and discriminability as primary diagnostics, targeting the two main failure modes of high-granularity refinement: granularity loss via distribution collapse and non-discriminative topic labels.

\subsection{Metrics}

\paragraph{Conventional metrics.}
We compute NPMI \cite{npmi}, $c_v$ \cite{cv}, and UMass \cite{umass} coherence using the Gensim Python library \cite{gensim}, as these remain widely adopted automated evaluation metrics for topic models despite known limitations in their alignment with human judgment \citep{coherenceinvalid1}. All coherence metrics and topic diversity \cite{diversity} are computed over the top-$10$ c-TF-IDF keywords per topic. Topic diversity is defined as the proportion of unique words across all topics. Sliding window sizes follow Gensim defaults ($110$ for $c_v$, $10$ for NPMI). All metrics use the target corpus as the reference text.

\paragraph{LLM-based evaluation.}
Following \citet{llmcoherence}, we use an LLM to approximate two
standard human evaluation tasks: (i)~\emph{coherence rating}, where
the model scores keyword relatedness on a 1--3 scale over the top-10
c-TF-IDF keywords, and (ii)~\emph{word intrusion}, where the model
identifies an intruder keyword injected among five topic keywords.
Word intrusion was introduced by \citet{chang2009} as a
human evaluation task that measures the semantic coherence of topics, specifically whether the representative words of the topics form a coherent enough group that the intruder word can be identified. \citet{llmcoherence} show that LLM
coherence ratings correlate more strongly with human judgments than
conventional automated metrics, and that LLM word-intrusion accuracy
approaches that of humans.  We use
GPT-4o~\cite{openai2024gpt4ocard} with domain-adapted prompts
(Appendix~\ref{app:prompts}).

\paragraph{Topic Dispersion.}
We quantify resistance to topic-size collapse using the effective number of topics, adapted from the effective number of species in ecology \citep{effectivenumber}, computed as the reciprocal of the Herfindahl--Hirschman Index. Let $p_k$ be the fraction of
documents assigned to topic $k$:
\begin{equation}
N_{\text{eff}} = \frac{1}{\sum_{k=1}^{K} p_k^2}
\end{equation}
To compare models with different $K$, we normalize to a 0--1 score:
\begin{equation}
\text{Dispersion} = \frac{N_{\text{eff}} - 1}{K - 1}
\end{equation}
A score of 1 indicates a uniform topic-size distribution; values near 0 indicate
collapse into a few dominant topics. The metric is most sensitive to concentration in a few dominant topics and less affected by size variation among smaller ones. We use dispersion as a primary outcome in Stage~B to quantify whether
refinement induces or prevents collapse.

\paragraph{Topic discriminability.}
We measure how well a topic representation distinguishes its assigned documents
from the broader corpus. Let $r_t$ be the embedding of a topic
representation (label + summary), and let $e_d$ be the embedding of a document
chunk $d$ under the same embedding model used for clustering. For each topic
$t$ with assigned set $D_t$, we compute mean cosine similarity to assigned
documents (intra) and to all non-outlier documents not assigned to that topic (inter):
\begin{equation}
\text{Intra}(t) = \frac{1}{|D_t|} \sum_{d \in D_t} \cos(r_t, e_d)
\end{equation}
\begin{equation}
\text{Inter}(t) = \frac{1}{|D \setminus D_t|} \sum_{d \notin D_t} \cos(r_t, e_d)
\end{equation}
and define discriminability as the margin:
\begin{equation}
\text{Discrim}(t) = \text{Intra}(t) - \text{Inter}(t)
\end{equation}
Discriminability is defined over label+summary representations produced during
refinement. Since refinement typically merges only a fraction of the total
topics, changes are too diluted to detect at the full topic-set level; we
therefore report the average over merged topics, where each merged
topic's inter-similarity is computed against all non-outlier documents outside
its own assignment set.

\section{Experiments}

\subsection{Experimental Setup}

\paragraph{Topic modeling pipeline.}
Sentence segmentation is performed using the "Segment Any Text" Python library
\cite{sat}\footnote{\url{https://huggingface.co/segment-any-text/sat-12l-sm}}
for sentence-level chunking. Document chunks are embedded with
Qwen3-Embedding-0.6B
\cite{qwen3embedding}\footnote{\url{https://huggingface.co/Qwen/Qwen3-Embedding-0.6B}}
using domain-specific instruction prompts (Appendix~\ref{app:prompts}).
GPT-4o is used for all LLM components across both stages:
representation initialization (from top-10 c-TF-IDF keywords and 5
representative documents), pairwise merge judgments, and evaluation
metrics.

\paragraph{Refinement hyperparameters.}
For EGIM (Algorithm~\ref{alg:egim}), we set the similarity threshold
$\tau=0.75$, target cluster size $s=15$, patience $N=3$, and a maximum
budget of 200 LLM calls. Each merge decision includes 5
representative documents per topic.

\paragraph{Preprocessing variants.}
Across both stages we evaluate six preprocessing variants: uncleaned
text; stopword removal; rule-based cleaning (regex); and three
LLM-assisted boilerplate-removal variants. Each LLM-assisted variant
clusters sentences in embedding space, samples representative
sentences, and uses an LLM to classify clusters as boilerplate for
removal. Full definitions and parameters are provided in
Appendix~\ref{app:preprocessing}.

\subsection{Baseline}

We compare EGIM against the iterative LLM-based topic reduction method
of \citet{janssens2025} (hereafter \textbf{LLM-Iter}). At each iteration, LLM-Iter presents the current topic representations to the LLM and prompts it to identify and merge the two most similar topics, repeating until a target count is reached. We use the same topic representation form as EGIM (LLM-generated label+summary) so that differences reflect the method structure rather than the topic representation type. Two key differences between methods remain: (i) EGIM uses embedding similarity to identify candidate pairs rather than relying on the LLM for candidate selection; and (ii) EGIM permits the LLM to decline a merge, whereas LLM-Iter merges unconditionally the candidates identified.

\paragraph{Matched final topic count.}
EGIM does not target a specific topic count, instead converging under a
fixed budget. For fair comparison, we first run EGIM on each base model
to obtain its converged count $K'$, then run LLM-Iter to the same
$K'$, ensuring both methods are evaluated for the same reduction degree.

\subsection{Calibration Sweep (Stage~A)}
\label{sec:stageA-results}

Stage~A identifies upstream pipeline settings suited to high-granularity
BERTopic, producing the base models for Stage~B.

\paragraph{Coherence metric reliability.}
Conventional coherence metrics show inconsistent behaviour across our corpora.  To assess reliability, we correlate each
conventional metric with the two LLM-based measures of
\citet{llmcoherence}: coherence rating, an extrinsic coherence measure
that does not rely on co-occurrence statistics of the underlying corpus,
and word-intrusion accuracy, a task-based metric where the model either
correctly identifies the intruder word or it does not.  UMass correlates
consistently with both LLM measures across all three corpora
($\rho=0.56$--$0.66$ with intrusion accuracy, $p<0.001$; surviving
Bonferroni correction for 18 comparisons).  NPMI and $c_v$ do not: both
exhibit sign reversals, correlating positively on \textsc{laser} but
negatively or non-significantly on \textsc{legal}
(Appendix Table~\ref{tab:metric-correlations}).  This aligns with prior
concerns about the reliability of automated coherence
metrics~\cite{coherenceinvalid1}.

\paragraph{Granularity selection.}
We target $K{=}320$ to enable the fine-grained topic distinctions
required by downstream applications such as competitive intelligence.
To verify that quality degrades gracefully at this granularity, we
compare across $K \in \{20, 40, 80, 160, 320\}$
(Table~\ref{tab:granularity}). Quality declines gradually overall,
with intrusion accuracy at $K{=}320$ still well above the 17\%
chance baseline.

\begin{table}[t]
  \centering
  \small
  \begin{tabular}{r c c c c}
    \toprule
    \textbf{K} & \textbf{UMass} & \textbf{Coh.} & \textbf{Intr.} & \textbf{Div.} \\
    \midrule
    20  & $-1.76$ & 2.34 & .66 & .89 \\
    40  & $-1.97$ & 2.36 & .67 & .85 \\
    80  & $-2.09$ & 2.32 & .64 & .80 \\
    160 & $-2.28$ & 2.26 & .62 & .76 \\
    320 & $-2.48$ & 2.19 & .59 & .72 \\
    \bottomrule
  \end{tabular}
  \caption{Topic quality across granularity levels, averaged over all
  configurations in the Stage~A sweep. Coh.\ = LLM coherence rating
  (1--3); Intr.\ = word-intrusion accuracy (chance = .17);
  Div.\ = topic diversity.}
  \label{tab:granularity}
\end{table}

\paragraph{Chunk size and preprocessing.}
At $K{=}320$, 2-sentence chunks perform best on two of three corpora
(Appendix Table~\ref{tab:chunk-k320}); we fix chunk size to 2 sentences
for Stage~B. The best preprocessing method differs by corpus (Appendix
Table~\ref{tab:preproc-k320}), so Stage~B retains all six variants,
yielding 18 base models (3 corpora $\times$ 6 methods).

\subsection{Post-hoc Refinement Results (Stage~B)}
\label{sec:stageB-results}
Since preprocessing variants within each corpus share underlying
documents, we report corpus-level aggregates ($n{=}3$) and emphasise
effect sizes and consistency.

\paragraph{Dispersion.}
Both methods perform the same number of merge operations to reach the
matched~$K'$ (mean~33). However, LLM-Iter concentrates these into
5.4~distinct target topics on average, whereas EGIM distributes merges
across 17.1~distinct targets. Under LLM-Iter, the same dominant topics
repeatedly absorb others, a process we refer to as \emph{semantic
snowballing}. Table~\ref{tab:dispersion} shows the resulting dispersion
gap: LLM-Iter preserves only 34\% to 88\% of original dispersion, with
severe collapse on SPAD, while EGIM maintains above 94\%.

\begin{table}[t]
  \centering
  \small
  \begin{tabular}{l ccc}
    \toprule
    & \textbf{LASER} & \textbf{LEGAL} & \textbf{SPAD} \\
    \midrule
    Original  & .533 & .355 & .369 \\
    EGIM      & .537\,{\scriptsize(101\%)} & .349\,{\scriptsize(98\%)} & .346\,{\scriptsize(94\%)} \\
    LLM-Iter  & .469\,{\scriptsize(88\%)} & .275\,{\scriptsize(78\%)} & .127\,{\scriptsize(34\%)} \\
    \bottomrule
  \end{tabular}
  \caption{Topic dispersion after refinement (higher is better).
  Percentages show relative change against the original model.
  EGIM wins 18/18 configurations.}
  \label{tab:dispersion}
\end{table}

\paragraph{Discriminability.}
Table~\ref{tab:discriminability} reports discriminability on merged
topics. EGIM consistently achieves higher scores across all three
corpora, with improvements ranging from +25\% (LASER) to +55\% (SPAD),
outperforming in 16 of 18 configurations (5/6~LASER, 5/6~LEGAL,
6/6~SPAD).

\begin{table}[t]
  \centering
  \small
  \begin{tabular}{lccc}
    \toprule
    \textbf{Corpus} & \textbf{EGIM} & \textbf{LLM-Iter} & \textbf{$\Delta$} \\
    \midrule
    LASER & .130 & .103 & +25\% \\
    LEGAL & .104 & .079 & +32\% \\
    SPAD  & .096 & .062 & +55\% \\
    \bottomrule
  \end{tabular}
  \caption{Topic discriminability on merged topics (higher is better).
  EGIM wins 16/18 configurations.}
  \label{tab:discriminability}
\end{table}

\paragraph{Semantic snowballing.}
Table~\ref{tab:snowball} illustrates semantic snowballing under LLM-Iter.
Six topics covering LIDAR, explosive detection, laser guidance, and
wavelength selection are progressively absorbed into a single topic.
EGIM evaluated several of the same pairs: \emph{LIDAR and LADAR
Systems} was paired with \emph{Weapon-Locating Ladar} (similarity
0.796), and \emph{Laser Range Finding} with \emph{Weapon-Locating
Ladar}. In both cases the LLM declined to merge. Across the full run, all six
topics remained separate under EGIM.
Further examples appear in Appendix~\ref{app:snowballing}.

\begin{table}[t]
  \centering
  \small
  \begin{tabular}{@{}l@{}}
    \toprule
    LIDAR and LADAR Systems \\
    \quad + Weapon-Locating Ladar \\
    \quad + Raman/Fluorescence Explosive Detection \\
    \quad + Laser Range Finding \\
    \quad + Semi-Active Laser Guidance \\
    \quad + Laser Wavelength Selection \\
    \midrule
    \textbf{Result:} Laser Targeting \& Communication \\
    \bottomrule
  \end{tabular}
  \caption{Semantic snowballing under LLM-Iter (LASER corpus).
  One topic absorbs five unrelated others.}
  \label{tab:snowball}
\end{table}

\begin{figure}[t]
  \centering
  \includegraphics[width=\columnwidth]{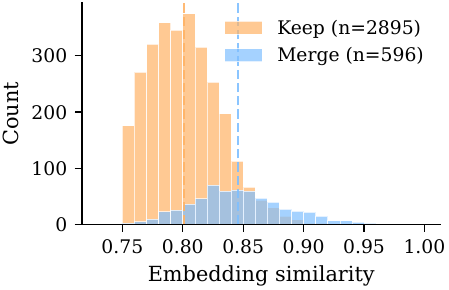}
  \caption{Embedding similarity (cosine) distributions for merged and declined
  pairs across 3{,}491 candidates. Dashed lines indicate class means.
  Heavy overlap confirms that no threshold reliably separates merge
  from decline decisions.}
  \label{fig:similarity}
\end{figure}

\paragraph{Limits of threshold-based merging.}
Across all runs, EGIM evaluated 3{,}491 candidate pairs above the
similarity threshold, merging 596 (17\%) and declining 2{,}895 (83\%).
Figure~\ref{fig:similarity} shows substantial overlap between the
similarity distributions of merged and declined pairs (mean 0.846
vs.\ 0.801), with overall AUC of 0.824 (per-corpus: 0.75 SPAD,
0.80 LEGAL, 0.89 LASER). The optimal threshold under Youden's~J
(0.823) catches only 72.7\% of true merges while wrongly merging
22.1\% of keeps. Raising the threshold to 0.85 reduces false merges
to 5.8\% but misses 57\% of LLM judged duplicates.

Table~\ref{tab:threshold} illustrates this:
\emph{Contract Interpretation Principles} and \emph{Statutory
Interpretation Principles} reach .893 similarity yet describe
fundamentally different legal doctrines, while \emph{Tax Assessment
Objections} and \emph{Taxation Objection Appeals} at .861 are redundant.

\begin{table}[t]
  \centering
  \small
  \begin{tabular}{@{}p{3cm}p{3cm}c@{}}
    \toprule
    \textbf{Topic A} & \textbf{Topic B} & \textbf{Sim} \\
    \midrule
    \multicolumn{3}{@{}l}{\textit{Declined (distinct concepts)}} \\
    \addlinespace[2pt]
    Contract Interp.\ Principles & Statutory Interp.\ Principles & .893 \\
    Misleading Conduct & Unconscionable Conduct & .854 \\
    \midrule
    \multicolumn{3}{@{}l}{\textit{Merged (true duplicates)}} \\
    \addlinespace[2pt]
    Tax Assessment Objections & Taxation Objection Appeals & .861 \\
    Extension of Time Applications & Extension of Time for Appeals & .875 \\
    \bottomrule
  \end{tabular}
  \caption{Similarity alone cannot determine merge decisions (LEGAL
  corpus). Distinct concepts may exceed the similarity of true
  duplicates.}
  \label{tab:threshold}
\end{table}

\paragraph{Overall model quality.}
Across both traditional ($c_v$, $c_{npmi}$, diversity) and
LLM-evaluated (coherence, intrusion) metrics, all changes remain
under 3\% for both methods (Appendix~\ref{app:stage-b-full}). This is expected: refinement affects only merged topics, which 
constitute roughly 10\% of the final topic set on average.

\paragraph{Convergence.}
The merge rate declines from approximately 30\% in the first three
rounds to below 8\% after round~8, aggregated over the 18
configurations. Some corpora exhaust the 200-call budget more often (SPAD: 6/6, 
LEGAL: 5/6, LASER: 2/6), reflecting more judged topic redundancy. Even in budget-limited runs,
productive merging is concentrated in early rounds; later calls
predominantly result in decline decisions. Convergence across iterations is visualised in Figure~\ref{fig:convergence} (Appendix~\ref{app:convergence}).

\paragraph{Coverage-based sampling ablation.}
To isolate the effect of k-means coverage gating, we compare EGIM
against a variant that selects candidate pairs by greedy similarity
ranking over the full topic space, holding all other components fixed
(single LASER configuration, equal LLM calls). Coverage gating
finds 56\% more merges (14 vs.\ 9) at 36\% better efficiency (12.6
vs.\ 19.7 calls per merge). Without coverage gating, candidate
selection concentrates on a narrower region of the topic space
(53.8\% of topics evaluated vs.\ 62.2\%), and individual topics
reappear more frequently (max 9 vs.\ 6 appearances), indicating
repeated evaluation of the same neighbourhood rather than
exploration of the full thematic space.

\paragraph{Computational cost.}
EGIM requires more LLM calls than LLM-Iter (mean 193.9 vs.\ 66.2,
a 2.9$\times$ increase). This reflects the cost of selectivity:
EGIM evaluates candidates that it ultimately declines, whereas
LLM-Iter merges unconditionally at each step. In practice the cost
is front-loaded, as the majority of productive merges occur in early
rounds with a steady decline in valid merges found as iterations
progress. A lower budget would capture most of the benefit.

\section{Conclusion}
We introduce EGIM, a methodology that can be applied to base topic
models for topic redundancy removal in high-granularity regimes.
Through a series of design decisions we resolve limitations of prior
post-hoc refinement strategies that result in incorrect merging and
distribution collapse, both leading to reduction in desired granularity.
To measure the common failure modes of high-granularity modelling we
proposed (i)~topic dispersion, a distribution metric that penalises
topic collapse, and (ii)~topic discriminability, a metric for measuring
the distinctiveness of a topic label within a topic model, key for
useful high-granularity topic modelling.

Across two patent corpora and a legal corpus, we found that 83\% of
high-similarity candidate pairs were rejected as non-redundant by LLM
judgment, confirming that semantic similarity alone cannot distinguish
true duplicates from distinct-but-related topics. EGIM distributed
merges across $3.1{\times}$ more distinct targets than the iterative
LLM baseline, preserving over 94\% of original dispersion while
improving discriminability on merged topics by 25--55\%.

\section*{Limitations}

\paragraph{Ablation Limitations.}
Our study utilised a single embedding model for core topic modelling
and topic representation sampling, and a single LLM for both the
methodology and evaluation. Whilst we chose a modern instruction-tuned
embedding model for our experiments, it would be interesting to
investigate how different model sizes and potential domain fine-tuning
alter results. In particular, our candidate selection uses
Qwen3-Embedding-0.6B; different embedding models may shift the similarity
distributions for the merge/keep decisions, and
domain-tuned embedding models may better
differentiate genuinely duplicate pairs from related-but-distinct
ones, potentially decreasing the AUC overlap of merge and keep
decisions when applied to specific corpora. The coverage-based sampling
ablation was also conducted on a single configuration rather than
across all~18, limiting the strength of that finding. Early testing
found that local LLM models struggled with producing high-quality
topic representations and topic evaluation, however we foresee this
limitation dissolving with the current upward performance trend of
smaller LLMs.

\paragraph{Evaluation Limitations.}
Stage~A hinted at an interesting result in conventional coherence
metric reliability at high granularity, with UMass maintaining
consistent correlation with our LLM human-task proxies. Whilst this
may be a result of the structure of the UMass metric (no sliding
windows, thus potentially handling rare vocabulary better), we did not
feel confident enough with our amount of data to make definitive
claims. In addition, whilst topic discriminability and topic dispersion
worked well to measure our targeted failure modes, we saw little
change in model-wide metrics after post-hoc refinement under either
method. We attribute this to the relatively low proportion of topics
that get altered in our high-granularity regime, but investigating how
these metrics change with more aggressive reduction (lowering the
similarity threshold, increasing the LLM call budget, or altering the
merge prompt to favour merge decisions more readily) would be an
interesting direction.

\paragraph{Corpora Limitations.}
Both our embedding model and LLM are capable of multilingual
processing, however all three corpora were constrained to English. It
would be a useful investigation to see how a multilingual dataset
alters the methodology's decisions in how topics are merged and
represented. In addition, we chose to use three specialised corpora
over traditional topic model benchmark corpora. This decision was made
because our corpora represent real-world targets for high-granularity
regimes that could provide valuable input to downstream use cases;
however, a trade-off in cross-work comparison was made, and future
investigation into post-hoc refinement for high-granularity regimes
should consider including standard benchmarks.

\paragraph{Closed-Source Models.}
Our methodology and LLM-based evaluation both rely on OpenAI
closed-source models, raising reproducibility concerns: model
versioning means our exact results may not be replicable if the
underlying model is updated or deprecated, and API costs present a
barrier to replication. Whilst we found early on that local models
gave inferior results for both topic representation and merge
judgment, a systematic comparison of local versus closed-source
performance would strengthen reproducibility and help establish the
minimum model capability required for effective merge decisions.

\section*{Acknowledgments}
This work was supported by the Engineering and Physical Sciences
Research Council (EPSRC), STMicroelectronics, and The Data Lab,
Edinburgh (Scottish Funding Council), through an industrial PhD
studentship at Heriot-Watt University.

\bibliography{custom}

%
%

\clearpage
\appendix
\raggedbottom


\section{Prompts}
\label{app:prompts}

All prompts used in the core pipeline are provided here.
Preprocessing-specific prompts are provided in
Appendix~\ref{app:preprocessing}. Bracketed placeholders (e.g.,
\texttt{[DOCUMENT]}) indicate dynamically substituted content. The
topic labelling and merge/keep prompts
(Prompts~\ref{prompt:initial-sys}--\ref{prompt:merge-keep}) include a
corpus-specific context line as the opening sentence of the system
message; the three corpora variants are shown in
Table~\ref{tab:corpus-context}. Evaluation prompts for LLM coherence
rating and word intrusion detection follow the protocol of
\citet{llmcoherence} with similar domain context.

\begin{table}[h]
  \centering
  \small
  \begin{tabular}{@{}lp{4.8cm}@{}}
    \toprule
    \textbf{Corpus} & \textbf{Context line} \\
    \midrule
    LASER & You are labeling topics from a corpus of laser technology patents filed by defense companies. \\
    \addlinespace[3pt]
    SPAD & You are labeling topics from a corpus of patents related to Single-Photon Avalanche Diode (SPAD) technology. \\
    \addlinespace[3pt]
    LEGAL & You are labeling topics from a corpus of Australian Federal Court legal cases. \\
    \bottomrule
  \end{tabular}
  \caption{Corpus-specific context lines injected into topic representation and merge/keep prompts.}
  \label{tab:corpus-context}
\end{table}

\noindent The placeholder \texttt{[DOMAIN FOCUS TERM]} is substituted
with ``technical focus'' for patent corpora and ``legal subject matter''
for the legal corpus. \texttt{[DOMAIN TERM]} is substituted with
``technology'' or ``legal concept'' respectively.

\subsection{Embedding Instruction}
\label{app:embedding-instruction}

Document chunks are embedded using Qwen3-Embedding-0.6B with the
task instruction shown in Prompt~\ref{prompt:embedding}.

\begin{prompt}[h]
\begin{promptbox}{Embedding Instruction}
\small
Instruct: Retrieve semantically similar text passages. Focus on novel
substantive content, key concepts, and unique information while
de-emphasizing formulaic or repetitive language.

Query: \texttt{[DOCUMENT CHUNK]}
\end{promptbox}
\caption{Task instruction for
  embedding with Qwen3-Embedding-0.6B.}
\label{prompt:embedding}
\end{prompt}

\subsection{Initial Topic Representation}
\label{app:initial-rep}

After BERTopic produces initial document--topic assignments and c-TF-IDF
keyword representations, we generate a descriptive label and summary for
each topic. The system message (Prompt~\ref{prompt:initial-sys})
defines output requirements; the user message
(Prompt~\ref{prompt:initial-user}) provides the top-10 c-TF-IDF
keywords and up to 10 representative documents selected via MMR-based
sampling (Appendix~\ref{app:egim-details}). Representative documents
are truncated to 800 characters each.

\begin{prompt*}[t]
\begin{promptbox}{Initial Topic Representation --- System Message}
\small
\texttt{[CORPUS CONTEXT]}
\medskip

Your task is to generate a concise label and summary for a topic based on its keywords and representative documents.
\medskip

CRITICAL REQUIREMENT:
\begin{itemize}[nosep,leftmargin=*]
  \item The label MUST accurately describe ALL representative documents, not just one or two
  \item First identify the COMMON THREAD that unifies all documents
  \item If documents show different applications of the same underlying technology/concept, label the underlying technology/concept
\end{itemize}
\medskip

LABEL GUIDELINES:
\begin{itemize}[nosep,leftmargin=*]
  \item Length: 2--6 words, Title Case
  \item Capture the specific \texttt{[DOMAIN FOCUS TERM]}
  \item Be specific enough to distinguish this topic from others
  \item Be general enough that it applies to ALL documents shown
  \item Avoid unexpanded acronyms unless universally known (e.g., ``LIDAR'' is OK, ``EMM'' is not)
  \item Prefer concrete technical terms over vague descriptors like ``Advanced'' or ``Novel''
\end{itemize}
\medskip

SUMMARY GUIDELINES:
\begin{itemize}[nosep,leftmargin=*]
  \item Single complete sentence, 100--180 characters
  \item Explain what the topic covers at an appropriate abstraction level
  \item Must be a complete sentence ending with a period
\end{itemize}
\medskip

OUTPUT FORMAT (JSON only, no explanation):\\
\texttt{\{"label": "<label>", "summary": "<summary>"\}}
\end{promptbox}
\caption{System message for initial topic representation. The LLM
  generates a label and summary from c-TF-IDF keywords and
  representative documents.}
\label{prompt:initial-sys}
\end{prompt*}

\begin{prompt*}[t]
\begin{promptbox}{Initial Topic Representation --- User Message}
\small
KEYWORDS: \texttt{[TOP-10 C-TF-IDF KEYWORDS]}
\medskip

REPRESENTATIVE DOCUMENTS:\\
{[Doc 1]}: \texttt{[PASSAGE 1]}\\
{[Doc 2]}: \texttt{[PASSAGE 2]}\\
\quad\vdots\\
{[Doc $N$]}: \texttt{[PASSAGE $N$]}
\medskip

Remember: Your label must apply to ALL documents above, not just the most distinctive one.
\medskip

OUTPUT:
\end{promptbox}
\caption{User message for initial topic representation, providing
  the c-TF-IDF keywords and up to 10 representative documents.}
\label{prompt:initial-user}
\end{prompt*}

\subsection{Merge/Keep Judgment}
\label{app:merge-keep}

For each candidate pair identified by embedding similarity, the LLM
receives both topic representations and their representative documents
(Prompt~\ref{prompt:merge-keep}) and makes a binary merge-or-keep
decision. Up to 5 representative documents per topic are included,
truncated to 200 characters each.

\begin{prompt*}[t]
\begin{promptbox}{Merge/Keep Judgment --- Prompt}
\small
\texttt{[CORPUS CONTEXT]}
\medskip

You must decide whether these two topics should be MERGED into one or KEPT as separate topics.
\medskip

TOPIC A: [ID: \texttt{[ID\_A]}]\\
Label: \texttt{[LABEL\_A]}\\
Summary: \texttt{[SUMMARY\_A]}\\
Representative documents:\\
\quad -- \texttt{[DOC\_A1]}\\
\quad -- \texttt{[DOC\_A2]}\\
\quad\quad\vdots
\medskip

TOPIC B: [ID: \texttt{[ID\_B]}]\\
Label: \texttt{[LABEL\_B]}\\
Summary: \texttt{[SUMMARY\_B]}\\
Representative documents:\\
\quad -- \texttt{[DOC\_B1]}\\
\quad -- \texttt{[DOC\_B2]}\\
\quad\quad\vdots
\medskip

\texttt{[DOMAIN-SPECIFIC MERGE GUIDANCE]}

\texttt{[DOMAIN-SPECIFIC KEEP GUIDANCE]}
\medskip

DECISION:
\begin{itemize}[nosep,leftmargin=*]
  \item If these topics are essentially the SAME \texttt{[DOMAIN TERM]} (even with different wording), choose MERGE
  \item If they are genuinely DIFFERENT \texttt{[DOMAIN TERM]}s, choose KEEP
\end{itemize}
\medskip

For MERGE: Specify which topic ID to keep and provide a new label/summary for the combined topic.\\
For KEEP: No further action needed.
\medskip

Respond with ONLY a JSON object:
\medskip

MERGE format:\\
\texttt{\{"action": "merge", "keep\_id": <id>, "new\_label": "<merged label>", "new\_summary": "<merged summary>"\}}
\medskip

KEEP format:\\
\texttt{\{"action": "keep"\}}
\end{promptbox}
\caption{Merge/keep judgment prompt. The LLM decides whether two
  candidate topics should be merged or kept separate based on their
  labels, summaries, and representative documents.}
\label{prompt:merge-keep}
\end{prompt*}

\noindent Domain-specific guidance is injected for merge and keep
decisions. Table~\ref{tab:merge-guidance} shows an example for the LASER
corpus; equivalent guidance is defined for SPAD and LEGAL corpora with
domain-appropriate terminology.

\begin{table*}[t]
  \centering
  \small
  \begin{tabular}{@{}l p{13cm}@{}}
    \toprule
    \textbf{Decision} & \textbf{Guidance (LASER corpus example)} \\
    \midrule
    MERGE &
    \textbullet\ Same core technology, even if applied differently
      (e.g., ``WDM Networks'' and ``WDM for Data Transmission'' $\rightarrow$ both WDM) \newline
    \textbullet\ One topic is a specific technique/variant of the other
      (e.g., ``Grating-Based Beam Combining'' is a type of ``Coherent Beam Combining'') \newline
    \textbullet\ Same physical principle with different implementations \newline
    \textbullet\ A domain expert would file these under one category for competitive analysis \\
    \addlinespace[6pt]
    KEEP &
    \textbullet\ Fundamentally different physical principles
      (e.g., ``Thermal Cooling'' vs ``Optical Sensing'') \newline
    \textbullet\ Completely different application domains
      (e.g., ``Medical Imaging'' vs ``Weapons Targeting'') \newline
    \textbullet\ Topics serve genuinely different analytical purposes \newline
    \textbullet\ A domain expert would want these tracked separately \\
    \bottomrule
  \end{tabular}
  \caption{Domain-specific merge/keep guidance injected into the judgment prompt (LASER corpus shown). SPAD and LEGAL corpora use equivalent guidance with domain-appropriate terminology and examples.}
  \label{tab:merge-guidance}
\end{table*}

\clearpage


\clearpage
\section{EGIM Implementation Details}
\label{app:egim-details}

This appendix item describes implementation details of the EGIM refinement
loop not covered by the pseudocode in Section~\ref{sec:method}.

\subsection{Representative Document Selection}
\label{app:doc-selection}

Each topic's representative text chunks are selected from the chunks
assigned to that cluster by HDBSCAN. Three progressively filtered
candidate sets are constructed:

\paragraph{Probability-gated set ($C_p$).}
Text chunks whose HDBSCAN cluster-assignment probability falls in the
top 20\% quantile for the topic.

\paragraph{Quality-gated set ($C_q \subseteq C_p$).}
Chunks from $C_p$ satisfying three criteria: (i)~minimum length of
350 characters, (ii)~at least 5\% of tokens match the topic's top
c-TF-IDF keywords, and (iii)~stopword density does not exceed 40\%.

\paragraph{Diversity-sampled set ($C_d \subseteq C_q$).}
Chunks selected from $C_q$ via Maximal Marginal Relevance
($\lambda{=}0.7$ relevance, $0.3$ diversity penalty). The first chunk
is the one closest to the $C_q$ centroid; subsequent chunks maximise the
MMR score.

\paragraph{Selection process.}
The target count is 10 chunks for initial labelling and 5 for merge/keep
judgments. $C_d$ forms the core of the representative set. If $C_q$ is
empty, $C_d$ is sampled from $C_p$ instead; if $C_p$ is also empty, all
cluster-assigned chunks are used. After MMR sampling, if fewer than the
target count have been selected (i.e.\ the sampled set was smaller than
the target), remaining slots are filled from all cluster-assigned chunks
ranked by assignment probability. This ensures the LLM always receives
sufficient context even for small clusters.

\subsection{Post-Merge Updates}
\label{app:post-merge}

When the LLM selects \textsc{merge}, three updates are applied to the
surviving topic before the next round begins:

\begin{enumerate}[nosep]
  \item \textbf{Document pooling.} Representative documents from both
    constituent topics are combined via round-robin interleaving and
    capped at 10. Since each list may itself be the product of earlier
    merges, this maintains thematic coverage across all constituent
    topics.
  \item \textbf{Re-embedding.} The merged label and summary are
    re-embedded with Qwen3-Embedding-0.6B so that subsequent similarity
    computations reflect the updated representation.
  \item \textbf{Cache invalidation.} All reviewed-pair entries
    involving either constituent topic ID are cleared, since the
    surviving topic's representation has changed and prior keep
    decisions may no longer hold.
\end{enumerate}


\clearpage
\section{Dataset Curation}
\label{app:datasets}

This appendix documents the retrieval and filtering steps applied to
each corpus. All patent data were obtained from
Lens.org~\citep{jefferson2021lens}.

\subsection{SPAD Patent Corpus}
\label{app:spad-corpus}

Patents were retrieved by CPC classification code, targeting
single-photon avalanche diode and related photodetector array
technologies. The inclusion codes were \texttt{H01L27/146*}
(photodiode arrays), \texttt{H10F30/225*} (avalanche photodiodes,
reclassified), and \texttt{H01L31/107*} (avalanche photodiodes,
legacy). Documents matching any of the following codes were excluded
to remove unrelated technologies sharing classification overlap:
\texttt{G01S13/93*}, \texttt{A47L9/28*}, \texttt{H04N5/232},
\texttt{E05D15*}, \texttt{H01L27/32*}, and \texttt{G02F1/1335}. Only
English-language granted patents and patent applications published
from January~2000 onward were included. Patent family deduplication
was applied. The search was executed in October~2024, yielding
1,102~documents. Each patent is represented by its description
section text.

\subsection{Laser Defense Patent Corpus}
\label{app:laser-corpus}

This corpus was constructed through per-assignee searches for
defense and aerospace organisations including: BAE~Systems, Elbit~Systems,
Israel Aerospace Industries~(IAI/Elta), L3Harris,
Lockheed~Martin, MBDA, Northrop~Grumman, Rafael Advanced Defense
Systems, Rheinmetall, RTX~(Raytheon), and Thales. For each
organisation, English-language patents were retrieved where the
applicant or owner name matched known variants of the company name and
at least one of the title, abstract, or claims fields contained one
or more of the following keywords: \texttt{laser*}, \texttt{LIDAR},
\texttt{"directed energy"}, \texttt{DIRCM},
\texttt{"optical parametric*"}, \texttt{"stimulated emission*"},
\texttt{"beam director*"}, and \texttt{"target illuminator*"}. No date
restriction was applied. Patent family deduplication was applied. The
final corpus contains 3,302~documents. Patents
are represented by their description section text.

\subsection{Legal Corpora}

The legal corpus consists of 3,890~case reports from the Federal Court
of Australia, sourced from AustLII and distributed via the UCI Machine
Learning Repository~\citep{legal_case_reports}. The dataset comprises
all available cases from 2006--2009. Each case is represented by its
full text.


\clearpage
\section{Optimisation Details}
\label{app:optimisation}

\subsection{Constrained Objective}
\label{app:objective}

We constrain the optimisation to prioritise feasibility: the TPE
sampler must first find configurations that produce at least
$K_{\text{target}}$ topics before it can meaningfully optimise for
quality. Trials where $n_{\text{topics}} < K_{\text{target}}$ are
marked infeasible via a normalised violation score
$v = \max(0,\; (K_{\text{target}} - n_{\text{topics}}) /
K_{\text{target}})$. Trials producing fewer than 2~topics or fewer than
100~non-outlier documents are pruned outright as degenerate.

Among feasible trials, the sampler maximises an equally weighted
composite of four components: $C_v$~coherence, topic diversity,
topic dispersion, and noise score ($1 - r_{\text{outlier}}$).

\subsection{Hyperparameter Search Space}
\label{app:search-space}

Table~\ref{tab:search-space} lists the tuned hyperparameters.

\begin{table}[h]
\centering
\small
\setlength{\tabcolsep}{4pt}
\begin{tabular}{@{}llll@{}}
\toprule
\textbf{Component} & \textbf{Parameter} & \textbf{Type} & \textbf{Range} \\
\midrule
UMAP    & \texttt{n\_neighbors}   & int  & $[5, 50]$ \\
        & \texttt{n\_components}   & cat & $\{5, 10, 15, 20\}$ \\
        & \texttt{min\_dist}       & float & $[0.0, 0.5]$ \\
\midrule
HDBSCAN & \texttt{min\_cluster\_size}       & int   & $[2, 200]$ \\
        & \texttt{min\_samples}              & int   & $[2, 60]$ \\
        & \texttt{cluster\_sel\_epsilon}    & float & $[0.0, 0.5]$ \\
\bottomrule
\end{tabular}
\caption{Hyperparameter search space for Stage~A optimisation.}
\label{tab:search-space}
\end{table}

\subsection{Optuna Configuration}
\label{app:optuna-config}

We use the TPE sampler with the constraint mechanism described above.
Each experiment runs 50~trials across target topic counts
$K \in \{20, 40, 80, 160, 320\}$. Dimensionality reduction and clustering both
use RAPIDS-accelerated GPU implementations (cuML UMAP and HDBSCAN).
Random seeds are fixed at~42 throughout.


\clearpage
\section{Preprocessing Definitions}
\label{app:preprocessing}

All six preprocessing variants are applied before embedding and clustering. We tested how removing irrelevant text (boilerplate) before running our topic modelling pipeline would interact with topic model performance and desired granularity.

\paragraph{Variant~1: Uncleaned.}
Raw text with no filtering.  Serves as the baseline.

\paragraph{Variant~2: Stopword removal.}
Standard NLTK English stopwords, in addition to corpus-specific
stopwords (tokens appearing in ${\geq}\,70\%$ of documents) are
removed at the token level.  This tests whether token-level
pre-embedding cleaning can improve subsequent text clustering and
thus topic modelling output.

\paragraph{Variant~3: Rule-based.}
Domain-specific regex patterns flag entire sentences as noise.  For
patents, six pattern groups target: section headings (``field of the
invention'', ``background of the disclosure''), filing metadata
(``claims the benefit of'', ``continuation of'', ``PCT application''),
figure and table references (``Fig.\ 3'', ``as illustrated in''),
bare specifications without context (``operating temperature'',
``frequency \ldots\ GHz''), generic computing scaffolding (``a
processor'', ``computer-readable medium''), and claim-style fragments
(``the method comprises'', ``the apparatus of'').  For legal
documents, patterns target procedural text (``appeal dismissed'',
``hearing date''), case citation references (``see \ldots\ v.'', ``in
re''), and structural headings.  Sentences shorter than 20~characters
or 4~tokens are also removed.

\paragraph{Variants~4--6: LLM-assisted semantic cleaning.}
These three variants share a common pipeline but differ in the
embedding model or instruction used for the initial sentence
clustering step for cleaning (not the chunk-level clustering that
precedes topic modelling).  The pipeline proceeds as follows:

\begin{enumerate}[nosep,leftmargin=*]
\item \textbf{Embed} all sentences in the corpus using the
  variant-specific model and instruction
  (Table~\ref{tab:llm-clean-variants}).
\item \textbf{Reduce} dimensionality with UMAP
  ($n_{\text{neighbors}}\!=\!15$, $n_{\text{components}}\!=\!30$,
  $\texttt{min\_dist}\!=\!0.1$).
\item \textbf{Cluster} the reduced embeddings with $K$-Means, where
  $K = \text{clamp}(\lfloor N/500 \rfloor,\; 50,\; 1000)$ and $N$ is
  the total sentence count.
\item \textbf{Sample} 10 diverse sentences per cluster via
  farthest-first traversal in embedding space.
\item \textbf{Classify} each cluster by prompting a local
  Gemma~3 27B\footnote{\citet{gemma}; \url{https://ollama.com/library/gemma3:27b}.} instance with the sampled sentences.
  For patents the model assigns one of five categories:
  \emph{specific\_technique}, \emph{generic\_capability},
  \emph{pure\_specification}, \emph{legal\_boilerplate}, or
  \emph{administrative}.  For legal documents:
  \emph{legal\_reasoning}, \emph{factual\_interpretation},
  \emph{statutory\_reference}, \emph{administrative\_procedural}, or
  \emph{structural\_transition}.
\item \textbf{Filter}: only sentences belonging to clusters labelled
  \emph{specific\_technique} (patents) or \emph{legal\_reasoning}
  (legal) are retained.
\end{enumerate}

\noindent The three variants differ solely in how sentences are
embedded before clustering, which changes which sentences end up
co-clustered and therefore which clusters the LLM sees:

\begin{table}[h]
\centering
\small
\setlength{\tabcolsep}{3pt}
\begin{tabular}{@{}cp{3.2cm}p{4.2cm}@{}}
\toprule
\textbf{Var.} & \textbf{Embedding Model} & \textbf{Instruction} \\
\midrule
4 & GTE-Multilingual-Base\footnotemark & None \\[3pt]
5 & Qwen3-Embedding-0.6B  & ``\ldots Emphasize identifying
  repetitive patterns, formulaic language, and structural
  boilerplate \ldots'' \\[3pt]
6 & Qwen3-Embedding-0.6B  & ``\ldots Focus on novel
  substantive content, key concepts, and unique
  information \ldots'' \\
\bottomrule
\end{tabular}
\caption{Embedding configurations for the three LLM-assisted
  cleaning variants.}
\label{tab:llm-clean-variants}
\end{table}

\footnotetext{\citet{gte}; \url{https://huggingface.co/Alibaba-NLP/gte-multilingual-base}.}

\noindent Variant~4, labelled \emph{LLM\,(GTE)}, uses a general-purpose encoder with
no task instruction.  Variants~5 and~6 use the same embedding model
employed in the main pipeline.  Variant~5, labelled \emph{LLM\,(noise)}, uses a task instruction that
attempts to focus the encoding on textual noise rather than signal.  Variant~6, labelled \emph{LLM\,(signal)}, uses the same
task instruction employed for topic modelling embeddings in the main
pipeline, focusing on substantive content.

\clearpage


\onecolumn
\section{Stage~A Supplementary Results}
\label{app:stageA}

This appendix provides the full tables summarised in
Section~\ref{sec:stageA-results}.

\vspace{1em}


\begin{center}
\small
\begin{tabular}{@{}l ccc ccc@{}}
\toprule
 & \multicolumn{3}{c}{--- LLM Coherence ---} &
   \multicolumn{3}{c}{--- LLM Intrusion ---} \\[2pt]
Metric & \textsc{laser} & \textsc{legal} & \textsc{spad}
       & \textsc{laser} & \textsc{legal} & \textsc{spad} \\
\midrule
$c_v$  & $+0.47^{***}$ & $-0.18$        & $+0.14$
       & $+0.58^{***}$ & $+0.03$        & $+0.60^{***}$ \\
NPMI   & $+0.55^{***}$ & $-0.34^{**}$   & $+0.23^{*}$
       & $+0.63^{***}$ & $-0.08$        & $+0.67^{***}$ \\
UMass  & $+0.57^{***}$ & $+0.65^{***}$  & $+0.58^{***}$
       & $+0.56^{***}$ & $+0.58^{***}$  & $+0.66^{***}$ \\[4pt]
LLM Intrusion & $+0.68^{***}$ & $+0.60^{***}$ & $+0.51^{***}$
              & --- & --- & --- \\
\bottomrule
\multicolumn{7}{@{}l}{\rule{0pt}{2.5ex}\footnotesize
  $*$\,$p{<}0.05$,\;\; $**$\,$p{<}0.01$,\;\; $***$\,$p{<}0.001$}
\end{tabular}
\captionof{table}{Spearman correlations ($\rho$) between conventional
  coherence metrics and the LLM-based measures of
  \citet{llmcoherence}.  Each correlation is computed over
  90 model-level observations per corpus (3~chunk sizes $\times$
  6~preprocessing methods $\times$ 5~granularity levels).  All
  correlations significant at $p{<}0.001$ survive Bonferroni
  correction for 18 comparisons ($\alpha_{\text{adj}} = 0.003$).
  The bottom row shows agreement between the two LLM-based measures.}
\label{tab:metric-correlations}
\end{center}

\vspace{1.5em}


\noindent
\begin{minipage}[t]{0.48\textwidth}
\centering
\small
\begin{tabular}{@{}ll rrrr@{}}
\toprule
\textbf{Corpus} & \textbf{Chunk} & \textbf{UMass} & \textbf{Coh.} & \textbf{Intr.} & \textbf{Div.} \\
\midrule
\multirow{3}{*}{\textsc{laser}}
  & 2-sent  & $\mathbf{-2.49}$ & $\mathbf{2.16}$ & $\mathbf{.67}$ & $\mathbf{.72}$ \\
  & 5-sent  & $-3.11$          & $2.08$          & $.62$          & $.70$ \\
  & 10-sent & $-2.97$          & $2.13$          & $.61$          & $.68$ \\
\midrule
\multirow{3}{*}{\textsc{legal}}
  & 2-sent  & $-2.41$          & $2.33$          & $.62$          & $\mathbf{.82}$ \\
  & 5-sent  & $\mathbf{-2.31}$ & $\mathbf{2.36}$ & $\mathbf{.63}$ & $.81$ \\
  & 10-sent & $-2.53$          & $2.32$          & $.57$          & $.80$ \\
\midrule
\multirow{3}{*}{\textsc{spad}}
  & 2-sent  & $\mathbf{-2.07}$ & $\mathbf{2.14}$ & $\mathbf{.56}$ & $.64$ \\
  & 5-sent  & $-2.14$          & $2.10$          & $.51$          & $\mathbf{.65}$ \\
  & 10-sent & $-2.23$          & $2.09$          & $.49$          & $.64$ \\
\bottomrule
\end{tabular}
\captionof{table}{Topic quality at $K{=}320$ by chunk size, averaged
  across preprocessing methods.  Bold = best per corpus.
  Coh.~= LLM coherence (1--3); Intr.~= word-intrusion accuracy
  (chance~= .17); Div.~= topic diversity.}
\label{tab:chunk-k320}
\end{minipage}%
\hfill
\begin{minipage}[t]{0.50\textwidth}
\centering
\small
\begin{tabular}{@{}ll rrrr@{}}
\toprule
\textbf{Corpus} & \textbf{Preproc.} & \textbf{UMass} & \textbf{Coh.} & \textbf{Intr.} & \textbf{Div.} \\
\midrule
\multirow{6}{*}{\textsc{laser}}
  & Uncleaned        & $-2.98$          & $2.09$          & $.64$          & $.71$ \\
  & Stopword         & $\mathbf{-2.61}$ & $2.09$          & $.63$          & $.69$ \\
  & Rule-based       & $-2.90$          & $2.08$          & $.63$          & $.71$ \\
  & LLM (GTE)        & $-2.87$          & $2.18$          & $.64$          & $.70$ \\
  & LLM (noise)      & $-3.07$          & $2.10$          & $.60$          & $.71$ \\
  & LLM (signal)       & $-2.70$          & $\mathbf{2.20}$ & $\mathbf{.65}$ & $.69$ \\
\midrule
\multirow{6}{*}{\textsc{legal}}
  & Uncleaned        & $-2.42$          & $2.36$          & $\mathbf{.64}$ & $.82$ \\
  & Stopword         & $\mathbf{-2.19}$ & $2.33$          & $.61$          & $.82$ \\
  & Rule-based       & $-2.51$          & $2.26$          & $.60$          & $\mathbf{.84}$ \\
  & LLM (GTE)        & $-2.20$          & $\mathbf{2.53}$ & $.62$          & $.76$ \\
  & LLM (noise)      & $-2.35$          & $2.36$          & $.62$          & $.79$ \\
  & LLM (signal)       & $-2.82$          & $2.18$          & $.52$          & $.81$ \\
\midrule
\multirow{6}{*}{\textsc{spad}}
  & Uncleaned        & $-2.09$          & $\mathbf{2.17}$ & $.53$          & $.63$ \\
  & Stopword         & $\mathbf{-1.83}$ & $2.01$          & $.52$          & $.64$ \\
  & Rule-based       & $-2.11$          & $2.12$          & $\mathbf{.56}$ & $\mathbf{.67}$ \\
  & LLM (GTE)        & $-2.40$          & $2.08$          & $.50$          & $.66$ \\
  & LLM (noise)      & $-2.17$          & $2.13$          & $.52$          & $.65$ \\
  & LLM (signal)       & $-2.24$          & $2.16$          & $.52$          & $.63$ \\
\bottomrule
\end{tabular}
\captionof{table}{Topic quality at $K{=}320$ by preprocessing method,
  averaged across chunk sizes.  Bold = best per corpus.  Preprocessing
  definitions in Appendix~\ref{app:preprocessing}.  Metric
  abbreviations as in Table~\ref{tab:chunk-k320}.}
\label{tab:preproc-k320}
\end{minipage}

\twocolumn


\onecolumn
\section{Stage~B Supplementary Results}
\label{app:stageB}


\subsection{Full Model Quality}
\label{app:stage-b-full}

Table~\ref{tab:full-quality} reports all quality metrics across methods
and corpora.  All grand-mean changes remain within $\pm$3\% of the
original model, confirming that refinement does not degrade overall
topic quality.

\begin{table}[h]
\centering
\small
\begin{tabular}{@{}ll rrr@{}}
\toprule
\textbf{Corpus} & \textbf{Metric}
  & \textbf{Original} & \textbf{EGIM} & \textbf{LLM-Iter} \\
\midrule
\multirow{6}{*}{\textsc{laser}}
  & $c_v$           & .710 & .710 & .710 \\
  & NPMI            & .154 & .156 & .155 \\
  & UMass           & $-2.46$ & $-2.46$ & $-2.47$ \\
  & LLM Coherence   & 2.38 & 2.40 & 2.38 \\
  & LLM Intrusion   & .701 & .716 & .703 \\
  & Diversity       & .724 & .731 & .730 \\
\midrule
\multirow{6}{*}{\textsc{legal}}
  & $c_v$           & .764 & .766 & .762 \\
  & NPMI            & .204 & .206 & .205 \\
  & UMass           & $-2.40$ & $-2.42$ & $-2.43$ \\
  & LLM Coherence   & 2.36 & 2.35 & 2.36 \\
  & LLM Intrusion   & .655 & .659 & .652 \\
  & Diversity       & .820 & .834 & .832 \\
\midrule
\multirow{6}{*}{\textsc{spad}}
  & $c_v$           & .671 & .677 & .678 \\
  & NPMI            & .125 & .123 & .124 \\
  & UMass           & $-1.92$ & $-2.01$ & $-2.00$ \\
  & LLM Coherence   & 2.18 & 2.20 & 2.16 \\
  & LLM Intrusion   & .582 & .597 & .602 \\
  & Diversity       & .645 & .678 & .682 \\
\midrule
\multirow{6}{*}{\textbf{Grand mean}}
  & $c_v$           & .715 & .718\,{\scriptsize(+0.3\%)} & .717\,{\scriptsize(+0.2\%)} \\
  & NPMI            & .161 & .162\,{\scriptsize(+0.4\%)} & .161\,{\scriptsize(+0.0\%)} \\
  & UMass           & $-2.26$ & $-2.30$\,{\scriptsize($-1.6$\%)} & $-2.30$\,{\scriptsize($-1.7$\%)} \\
  & LLM Coherence   & 2.30 & 2.31\,{\scriptsize(+0.5\%)} & 2.30\,{\scriptsize($-0.1$\%)} \\
  & LLM Intrusion   & .646 & .657\,{\scriptsize(+1.8\%)} & .652\,{\scriptsize(+1.0\%)} \\
  & Diversity       & .730 & .747\,{\scriptsize(+2.4\%)} & .748\,{\scriptsize(+2.5\%)} \\
\bottomrule
\end{tabular}
\caption{Full model quality across all metrics and methods.
  Corpus rows show means over 6~preprocessing variants;
  grand-mean rows average the three corpus means.
  Parenthesised percentages show change relative to the original model.
  All changes remain within $\pm$3\%.}
\label{tab:full-quality}
\end{table}


\subsection{Convergence Behaviour}
\label{app:convergence}

\begin{figure}[h]
  \centering
  \includegraphics[width=0.5\textwidth]{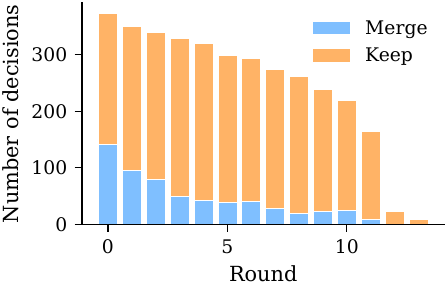}
  \caption{Merge and decline decisions across refinement rounds,
  aggregated over 18 configurations.  Productive merging is
  concentrated in early rounds; later rounds are dominated by
  decline decisions.}
  \label{fig:convergence}
\end{figure}

Figure~\ref{fig:convergence} shows merge and decline decisions across
refinement rounds.  The merge rate declines from approximately 30\%
in the first three rounds to below 8\% after round~8, confirming that
the algorithm converges toward a stable topic set.  Corpora with
denser similarity structure exhaust the 200-call budget more often
(SPAD: 6/6, LEGAL: 5/6, LASER: 2/6), reflecting a larger pool of
candidate pairs.


\subsection{Per-Corpus Similarity Distributions}
\label{app:per-corpus-sim}

\begin{figure}[h]
  \centering
  \includegraphics[width=\textwidth]{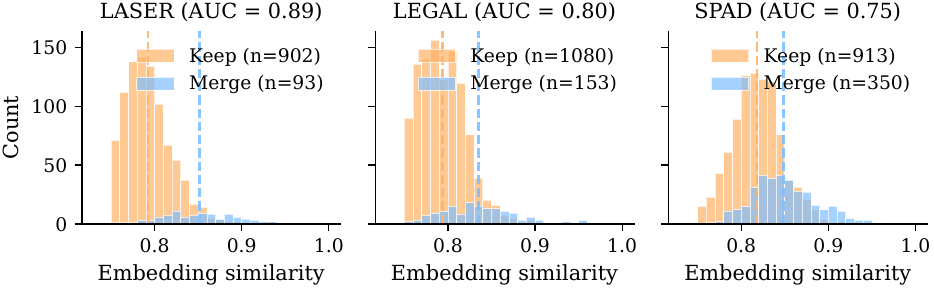}
  \caption{Embedding similarity distributions for merged and declined
  pairs, separated by corpus.  Dashed lines indicate class means.
  LASER shows the cleanest separation (AUC\,=\,0.89); SPAD's denser
  similarity structure produces heavy overlap (AUC\,=\,0.75).}
  \label{fig:sim-per-corpus}
\end{figure}

The aggregate AUC of 0.824 reported in
Section~\ref{sec:stageB-results} masks substantial corpus-level
variation.  Figure~\ref{fig:sim-per-corpus} shows that LASER's
sparser topic space produces well-separated merge and decline
distributions, while SPAD's smaller corpus yields inherently more
similar topics, compressing both distributions into a narrower
similarity band.  This confirms that similarity alone is insufficient
for merge decisions, particularly in dense corpora.

\clearpage
\subsection{Semantic Snowballing Examples}
\label{app:snowballing}

Figures~\ref{fig:snowball-optical}--\ref{fig:snowball-spad} show
snowball chains observed under LLM-Iter, where successive merges
broaden a topic until it absorbs unrelated neighbours.
Tables~\ref{tab:egim-optical}--\ref{tab:egim-spad} show EGIM's
decisions on the same topics.

\clearpage


\begin{figure*}[b]
\small
\setlength{\fboxsep}{6pt}
\noindent\fbox{%
\begin{minipage}{0.96\textwidth}
\textbf{LLM-Iter chain: 6 topics $\rightarrow$ 1}

\medskip\noindent
\textbf{Base topic:} Narrow Bandpass Optical Filters\\[2pt]
\textit{``\dots design and application of narrow bandpass optical
  filters that selectively transmit specific wavelength ranges
  while blocking others.''}

\medskip\noindent
\hspace{1em}$\oplus$~Diffraction Grating Systems\\
\hspace{1em}$\rightarrow$~\textbf{Optical Filtering and Dispersion Systems}\\[2pt]
\textit{``Covers optical filters and diffraction gratings used to
  selectively transmit wavelengths and control beam dispersion in
  laser systems.''}

\medskip\noindent
\hspace{1em}$\oplus$~Optical Comb Filtering and Sampling\\
\hspace{1em}$\rightarrow$~\textbf{Optical Filtering and Signal Processing}\\[2pt]
\textit{``Includes optical filters, diffraction gratings, comb filters,
  and samplers for wavelength selection, beam dispersion, and frequency
  analysis in laser systems.''}

\medskip\noindent
\hspace{1em}$\oplus$~Optical Reflection and Refraction\\
\hspace{1em}$\rightarrow$~\textbf{Optical Filtering and Beam Manipulation}\\[2pt]
\textit{``Covers optical filters, diffraction gratings, mirrors, and
  prisms for wavelength selection, beam dispersion, reflection, and
  refraction in laser systems.''}

\medskip\noindent
\hspace{1em}$\oplus$~Beam Expansion and Collimation\\
\hspace{1em}$\rightarrow$~\textbf{Optical Beam Manipulation Techniques}\\[2pt]
\textit{``Covers optical filters, diffraction gratings, mirrors,
  prisms, and lens arrangements for wavelength selection, beam
  dispersion, expansion, and collimation in laser systems.''}

\medskip\noindent
\hspace{1em}$\oplus$~Optical Beam Splitters\\
\hspace{1em}$\rightarrow$~\textbf{Optical Beam Manipulation and Splitting}\\[2pt]
\textit{``Covers optical systems for beam manipulation including
  filtering, splitting, expansion, collimation, and shaping.''}

\end{minipage}}
\caption{LLM-Iter snowball chain (\textsc{laser} corpus).
  Six distinct optical components collapse into a generic
  ``beam manipulation'' topic.}
\label{fig:snowball-optical}
\end{figure*}

\begin{table*}[t]
\footnotesize
\renewcommand{\arraystretch}{1.15}
\begin{tabular}{@{}p{0.42\textwidth} p{0.42\textwidth} c c@{}}
\toprule
\textbf{Topic~A (label + summary)} & \textbf{Topic~B (label + summary)} & \textbf{Sim.} & \textbf{Dec.} \\
\midrule

\textbf{Beam Expansion and Collimation.\textsuperscript{$\star$}}
Use of multi-lens arrangements and beam expanders to manipulate laser beam divergence and collimation.
&
\textbf{Laser Beam Expanders with Liquid Lenses.}
Beam expanders utilising liquid lens elements, often made with Cargille liquids, to achieve specific optical properties.
& .860 & \textsc{merge} \\

\midrule

\textbf{Optical Reflection and Refraction.\textsuperscript{$\star$}}
Use of mirrors and prisms for manipulating laser beams through reflection and refraction in optical systems.
&
\textbf{Turning Mirror in Laser Cladding.}
Use of a turning mirror in laser cladding systems to redirect beams and protect components during welding operations.
& .838 & \textsc{decline} \\[4pt]

\textbf{Optical Beam Splitters.\textsuperscript{$\star$}}
Use of beam splitters to divide laser beams into multiple segments for various applications, including measurement and signal processing.
&
\textbf{Optical Signal Distribution in Sensor Arrays.}
Methods and systems for distributing optical signals to sensor arrays using couplers, distribution buses, and return buses.
& .799 & \textsc{decline} \\[4pt]

\textbf{Narrow Bandpass Optical Filters.\textsuperscript{$\star$}}
Narrow bandpass filters designed to selectively transmit specific wavelengths while blocking others, often used in laser applications.
&
\textbf{Metamaterial Filters in IR Systems.}
Use of metamaterial filters in active infrared systems for spectral filtering and handling various laser wavelengths and incident angles.
& .789 & \textsc{decline} \\[4pt]

\textbf{Optical Comb Filtering and Sampling.\textsuperscript{$\star$}}
Use of optical comb filters and samplers to process and integrate modulated spectral copies for frequency analysis and signal processing.
&
\textbf{Optical and Femtosecond Frequency Combs.}
Generation, stabilisation, and application of optical frequency combs, consisting of multiple phase-locked frequency tones.
& .787 & \textsc{decline} \\[4pt]

\textbf{Diffraction Grating Systems.\textsuperscript{$\star$}}
Use of diffraction gratings to manipulate and control beam dispersion and spatial chirping in optical systems.
&
\textbf{In-Plane and Bevel Angles in Waveguides.}
Manipulation of in-plane and bevel angles in optical waveguides to control photon trajectories and enhance light guidance.
& .779 & \textsc{decline} \\

\bottomrule
\end{tabular}
\caption{EGIM decisions on the six chain topics from
  Figure~\ref{fig:snowball-optical} (\textsc{laser} corpus).
  Each starred ($\star$) topic is paired with its
  highest-similarity partner under EGIM; none of the six are
  paired with each other, and all are preserved as distinct.}
\label{tab:egim-optical}
\end{table*}


\clearpage

\begin{figure*}[b]
\small
\setlength{\fboxsep}{6pt}
\noindent\fbox{%
\begin{minipage}{0.96\textwidth}
\textbf{LLM-Iter chain: 3 topics $\rightarrow$ 1}

\medskip\noindent
\textbf{Base topic:} Copyright Infringement Analysis\\[2pt]
\textit{``\dots legal analysis of copyright infringement, focusing on
  the concepts of substantial similarity, originality, and the
  reproduction of works.''}

\medskip\noindent
\hspace{1em}$\oplus$~Copyright Infringement Dispute\\
\hspace{1em}$\rightarrow$~\textbf{Copyright Infringement Cases}\\[2pt]
\textit{``Encompasses legal cases and analyses related to copyright
  infringement, including the examination of substantial similarity,
  originality, reproduction of works, and specific disputes such as
  architectural plans.''}

\medskip\noindent
\hspace{1em}$\oplus$~Damages for IP Infringement\\
\hspace{1em}$\rightarrow$~\textbf{Copyright and IP Infringement Cases}\\[2pt]
\textit{``Covers legal cases and analyses related to copyright and
  intellectual property infringement, including the examination of
  substantial similarity, originality, reproduction of works, and the
  assessment and awarding of damages.''}

\end{minipage}}
\caption{LLM-Iter snowball chain (\textsc{legal} corpus).
  A doctrinal analysis topic and a case-specific dispute are
  reasonably merged, but then a distinct remedies topic
  (damages assessment) is absorbed, conflating
  infringement analysis with damages quantification.}
\label{fig:snowball-legal}
\end{figure*}

\begin{table*}[t]
\footnotesize
\renewcommand{\arraystretch}{1.15}
\begin{tabular}{@{}p{0.42\textwidth} p{0.42\textwidth} c c@{}}
\toprule
\textbf{Topic~A (label + summary)} & \textbf{Topic~B (label + summary)} & \textbf{Sim.} & \textbf{Dec.} \\
\midrule

\textbf{Copyright Infringement Analysis.\textsuperscript{$\star$}}
Legal analysis of copyright infringement, focusing on the concepts of substantial similarity, originality, and the reproduction of works.
&
\textbf{Copyright Infringement Dispute.\textsuperscript{$\star$}}
A legal dispute involving allegations of copyright infringement related to architectural plans created by Atrium and used by the Gaunt brothers and Beaumonde.
& .805 & \textsc{merge} \\[4pt]

\multicolumn{4}{@{}p{0.92\textwidth}}{%
\textbf{Damages for IP Infringement\textsuperscript{$\star$}}:
not selected for any merge by EGIM; preserved as a distinct topic.} \\

\bottomrule
\end{tabular}
\caption{EGIM decisions on the three chain topics from
  Figure~\ref{fig:snowball-legal} (\textsc{legal} corpus).
  EGIM merges the two infringement-analysis topics
  but preserves the damages/remedies topic that
  LLM-Iter absorbed.}
\label{tab:egim-legal}
\end{table*}


\clearpage

\begin{figure*}[b]
\small
\setlength{\fboxsep}{6pt}
\noindent\fbox{%
\begin{minipage}{0.96\textwidth}
\textbf{LLM-Iter chain: 4 topics $\rightarrow$ 1}

\medskip\noindent
\textbf{Base topic:} Photodiode Array Structures\\[2pt]
\textit{``\dots design and arrangement of photodiode arrays, including
  avalanche photodiodes, within semiconductor substrates for light
  detection.''}

\medskip\noindent
\hspace{1em}$\oplus$~Photodiode Array Configurations\\
\hspace{1em}$\rightarrow$~\textbf{Photodiode Array Design and Configurations}\\[2pt]
\textit{``Encompasses the design, arrangement, and configurations of
  photodiode arrays, including avalanche photodiodes, within
  semiconductor substrates for light detection and precise photon
  counting.''}

\medskip\noindent
\hspace{1em}$\oplus$~Photodetector and Semiconductor Substrate Integration\\
\hspace{1em}$\rightarrow$~\textbf{Photodetector Array Design and Integration}\\[2pt]
\textit{``Covers the design, arrangement, and integration of
  photodetector arrays, including avalanche photodiodes, with
  semiconductor substrates.  Includes methods of manufacture and
  device configurations.''}

\medskip\noindent
\hspace{1em}$\oplus$~HA and VA Photodetector Arrays\\
\hspace{1em}$\rightarrow$~\textbf{Photodetector Array Design and Configuration}\\[2pt]
\textit{``Encompasses the design, arrangement, and integration of
  photodetector arrays, including avalanche photodiodes, with
  semiconductor substrates.  Covers various configurations such as
  horizontal and vertical arrays.''}

\end{minipage}}
\caption{LLM-Iter snowball chain (\textsc{spad} corpus).
  An array-structures topic absorbs substrate-integration
  methods and specific array geometries (HA/VA),
  producing a generic ``design and configuration'' catch-all.}
\label{fig:snowball-spad}
\end{figure*}

\begin{table*}[t]
\footnotesize
\renewcommand{\arraystretch}{1.15}
\begin{tabular}{@{}p{0.42\textwidth} p{0.42\textwidth} c c@{}}
\toprule
\textbf{Topic~A (label + summary)} & \textbf{Topic~B (label + summary)} & \textbf{Sim.} & \textbf{Dec.} \\
\midrule

\textbf{Photodiode Array Structures.\textsuperscript{$\star$}}
Design and arrangement of photodiode arrays, including avalanche
photodiodes, within semiconductor substrates for light detection.
&
\textbf{Photodiode Array Configurations.\textsuperscript{$\star$}}
Various configurations and applications of photodiode arrays,
including avalanche photodiodes, for precise photon detection.
& .907 & \textsc{merge} \\

\midrule

\multicolumn{4}{@{}p{0.92\textwidth}}{%
\textbf{Photodetector--Substrate Integration\textsuperscript{$\star$}}
and \textbf{HA and VA Photodetector Arrays\textsuperscript{$\star$}}:
not selected for any merge by EGIM; preserved as distinct topics.} \\

\bottomrule
\end{tabular}
\caption{EGIM decisions on the four chain topics from
  Figure~\ref{fig:snowball-spad} (\textsc{spad} corpus).
  EGIM merges the two near-duplicate array topics
  but preserves the substrate-integration topic and the
  specific HA/VA geometry topic that LLM-Iter absorbed.}
\label{tab:egim-spad}
\end{table*}

\twocolumn

\end{document}